\def\year{2022}\relax
\documentclass[letterpaper]{article} 
\usepackage{aaai22}  
\usepackage{times}  
\usepackage{helvet}  
\usepackage{courier}  
\usepackage[hyphens]{url}  
\usepackage{graphicx} 
\usepackage{caption} 
\DeclareCaptionStyle{ruled}{labelfont=normalfont,labelsep=colon,strut=off} 
\usepackage{algorithm}
\usepackage{algorithmic}

\usepackage{newfloat}
\usepackage{listings}
\floatstyle{ruled}
\newfloat{listing}{tb}{lst}{}
\floatname{listing}{Listing}
\usepackage{amsmath}
\usepackage{hyperref}

\title{Spline-PINN: Approaching PDEs without Data \\ using Fast, Physics-Informed Hermite-Spline CNNs}
\author{Anonymous submission (5263)
}
\author {
    Nils Wandel,\textsuperscript{\rm 1}
    Michael Weinmann,\textsuperscript{\rm 2}
    Michael Neidlin,\textsuperscript{\rm 3}
    Reinhard Klein \textsuperscript{\rm 1}
}
\affiliations {
    \textsuperscript{\rm 1} University of Bonn\\
    \textsuperscript{\rm 2} Delft University of Technology\\
    \textsuperscript{\rm 3} RWTH Aachen University\\
    wandeln@cs.uni-bonn.de, m.weinmann@tudelft.nl, neidlin@ame.rwth-aachen.de, rk@cs.uni-bonn.de
}

\usepackage{color}      
\usepackage{comment}    
\begin{document}

\maketitle

\begin{abstract}





Partial Differential Equations (PDEs) are notoriously difficult to solve. In general, closed-form solutions are not available and numerical approximation schemes are computationally expensive. In this paper, we propose to approach the solution of PDEs based on a novel technique that combines the advantages of two recently emerging machine learning based approaches. 
%
%
First, physics-informed neural networks (PINNs) learn continuous solutions of PDEs and can be trained with little to no ground truth data. However, PINNs do not generalize well to unseen domains. Second, convolutional neural networks provide fast inference and generalize but either require large amounts of training data or a physics-constrained loss based on finite differences that can lead to inaccuracies and discretization artifacts.

We leverage the advantages of both of these approaches by using Hermite spline kernels in order to continuously interpolate a grid-based state representation that can be handled by a CNN. This allows for training without any precomputed training data using a physics-informed loss function only and provides fast, continuous solutions that generalize to unseen domains.
%
%
We demonstrate the potential of our method at the examples of the incompressible Navier-Stokes equation and the damped wave equation. Our models are able to learn several intriguing phenomena such as Karman vortex streets, the Magnus effect, Doppler effect, interference patterns and wave reflections. Our quantitative assessment and an interactive real-time demo show that we are narrowing the gap in accuracy of unsupervised ML based methods to industrial solvers for computational fluid dynamics (CFD) while being orders of magnitude faster. 




\end{abstract}

\section{Introduction}
Partial differential equations (PDEs) are an important mathematical concept to describe for example the motion of fluids, the propagation of waves, the evolution of stock markets, gravity and more. 
However, solving partial differential equations is a hard problem since closed-form solutions are rarely available. 
Thus, developing fast and accurate numerical schemes in order to find approximate solutions is of great interest for 
applications such as e.g. physics engines in computer games, computer generated imagery (CGI) for movies, or computational fluid dynamics (CFD) to help engineers with simulated wind tunnel experiments.

Recently, advances of machine learning (ML) based approaches have led to promising results coping with the high computational costs associated with classical numerical methods. 
Furthermore, some ML based approaches are by design differentiable. This means that they offer gradients that can be used for stability analysis, optimal control or reinforcement learning. 
However, ML based approaches often do not generalize to domains not seen during training \cite{Kim:2020,mohan2020embedding,thuerey2019deep,um2020solverintheloop,Raissi:2019} or rely on large amounts of training data which capture the variations to be expected in foreseen scenarios \cite{pfaff2021learning,kim2019deep,Ladicky:2015}.
%
Recent physics-constrained approaches based on finite differences  \cite{tompson2017accelerating,Zhu:2019,Wandel:2020,wandel-2020-3Dfluid} exhibit the potential to mitigate these problems but might lead to inaccuracies and discretization artifacts, especially at high Reynolds-Numbers. 

Here, we propose to use a Hermite spline CNN to obtain continuous solutions that can be trained with a physics-informed loss only. 
This approach combines the advantages of i) leveraging physics-informed neural networks (PINNs) to overcome the need for large amounts of training data \cite{Jin_2021,raissi2018hidden,Raissi:2019} and ii) leveraging the faster inference and better generalization capabilities offered by convolutional neural networks \cite{tompson2017accelerating,Wandel:2020,wandel-2020-3Dfluid}. 
%
We demonstrate the effectiveness of our approach for the incompressible Navier-Stokes equations as well as the damped wave equation. The incompressible Navier-Stokes equations, which are particularly hard to solve due to the non-linear advection term and thus the main focus of this paper, are investigated for Reynolds numbers ranging from 2-10000. To assess the accuracy of our method, we compute drag and lift coefficients on a CFD benchmark domain~\cite{cfd-benchmark} and compare the results with official benchmark values. For both equations, we perform generalization experiments, evaluate the stability over long time horizons and present an interactive real-time demonstration. 
To ensure full reproducibility, our code is publicly available on github: \url{https://github.com/aschethor/Spline_PINN}. 

\section{Related Work}

Recent developments indicate the potential of efficient surrogate models based on machine learning, and in particular deep learning, to approximate the dynamics of partial differential equations (PDEs). 
%

%
\paragraph{Lagrangian methods} like smoothed particle hydrodynamics (SPH)~\cite{gingold1977smoothed} represent fluids in the reference frame of individual particles that follow the fluid's velocity field.
%
Relying on this principle, learning-based Lagrangian approaches include the simulation of fluids based on regression forests~\cite{Ladicky:2015}, graph neural networks~\cite{Mrowca:2018,li2019learning}, continuous convolutions~ \cite{Ummenhofer:2020} and Smooth Particle Networks (SP-Nets)~ \cite{schenck2018spnets}.
%
%
While Lagrangian methods are particularly suitable for fluid domains that exhibit large, dynamic surfaces such as waves and droplets, however, Eulerian methods typically allow a more accurate simulation of fluid dynamics within a certain fluid domain (see~ \cite{Foster:1996}).
%

%
\paragraph{Eulerian methods} model domain characteristics such as a velocity or pressure field on a fixed reference frame. 
Respective techniques leverage implicit neural representations, grid- and mesh-structures to describe the domain of a PDE. 
%

\emph{Continuous Eulerian methods} exploit the direct mapping of domain coordinates (i.e. positional coordinates and time) to field values (i.e. velocity, pressure, etc.) relying e.g. on implicit neural representations to obtain smooth and accurate simulations~ \cite{Sirignano:2018,grohs2018proof,khoo2019solving,Raissi:2019,deepxde2021} and handle the curse of dimensionality faced by discrete techniques in the case of high-dimensional PDEs~\cite{grohs2018proof}. 
Respective applications include the modeling of flow through porous media~ \cite{Zhu:2018,Zhu:2019,Tripathy:2018}, fluid modeling~\cite{Yang:2016,raissi2018hidden}, turbulence modeling~\cite{Geneva:2019,ling2016reynolds}, wave propagation \cite{rashtbehesht2021physicsinformed,sitzmann2020implicit} and the modeling of molecular dynamics ~\cite{Schoeberl:2019}. 
Training typically relies on penalizing residuals of the underlying PDEs based on physics-informed loss functions for a specific domain, which prevents the networks from generalizing to new domain geometries and being used in the scope of interactive scenarios. Recently, Wang et al.~\cite{wang2021trainonce} proposed to stitch pretrained Deep Learning models together to speed up the learning process for boundary value problems. However, temporal dependencies are not embedded in their method yet preventing interactive applications.

\emph{Discrete Eulerian methods} such as finite difference, lattice Boltzmann~\cite{chen1998lattice,doi:10.1063/5.0041446} or finite element methods instead tackle the underlying PDEs on an explicit representation such as a discrete grid or a graph structure. 
%
Beyond early seminal work~\cite{harlow1965numerical,stam1999stable}, recent developments focus on exploiting the potential of deep learning techniques to achieve significant speed-ups while preserving accuracy. 
%
Learning fluid simulations has been approached with deep generative models for efficient interpolations between various simulation settings~\cite{kim2019deep} and turbulent flow fields within pipe domains~\cite{Kim:2020}. 
%
%
However, both of these approaches do not generalize to new domain geometries that have not been used during training. 
%
%
%
To speed up Eulerian fluid simulations, Tompson et al.~\cite{tompson2017accelerating} proposed to learn a Helmholtz projection step with a CNN that generalizes to new domain geometries that have not been used during the training. 
%
However, this approach requires a path tracer to deal with the advection term of the Navier-Stokes equations. Furthermore, as viscosity is not considered in Eulerian fluid models, phenomena such as the Magnus effect and Karman vortex streets cannot be simulated.
%
Ensuring incompressibility of fluids has been achieved with discretized vector potentials~\cite{kim2019deep,mohan2020embedding}, however, these techniques are not capable of generalizing to domain geometries not seen during the training. 
The approach by Geneva et al.~\cite{geneva2020modeling} discards the pressure term in the Navier-Stokes equations which leads to the simpler Burgers’ equations, for which they learned the update step based on a physics-constrained framework. 
Thuerey et al.~\cite{thuerey2019deep} learn to solve the Reynolds-averaged Navier-Stokes equations, but the specific approach prevents a generalization beyond airfoil flows and discards the temporal evolution of the fluid state. 
%
%
Um et al.~\cite{um2020solverintheloop} proposed to learn a correction step which allows approximating solutions of a high-resolution fluid simulations in terms of a low-resolution differentiable fluid solver, but the generalization capability to domain geometries that have not been used during training has not been shown. 
%
%
In contrast to the aforementioned techniques, the approach by Wandel et al.~\cite{Wandel:2020} and its 3D extension~\cite{wandel-2020-3Dfluid} overcome the dependence on huge amounts of training data by using a physics-constrained loss and introducing a training cycle that recycles data generated by the network during training. This allows handling dynamic boundary conditions as required for interactions with the fluid ~\cite{Wandel:2020,wandel-2020-3Dfluid} and, with an improved network~\cite{wandel-2020-3Dfluid}, also dealing with temporally varying fluid parameters such as viscosity and density during simulation. However, their physics-constrained loss can lead to discretization artifacts and requires a pressure gradient regularizer for high Reynolds numbers. 
%
%
%
Further learning-based approaches also exploit graph representations in terms of graph neural networks~\cite{harsch2021direct,sanchezgonzalez2020learning}, graph convolutional networks \cite{gao2021pigngn} as well as mesh representations~\cite{pfaff2021learning} or subspace representations~\cite{sirignano2018dgm,ainsworth2021galerkin}. 
%
Unfortunately, graph neural networks cannot make use of the highly efficient implementations for convolutional operations on grids and thus usually come with higher computational complexity per node compared to grid based CNNs. 
%

%
To the best of our knowledge, the potential of Spline CNNs ~\cite{Fey2017SplineCNN} has not yet been investigated for learning the dynamics of PDEs and, hence, the presented physics-informed Hermite Spline CNN (Spline-PINN) approach is the first method that leverages both implicit and explicit characteristics to obtain 
fast, interactive, continuous surrogate models that generalize and can be trained without training data using a physics-informed loss based on the underlying PDEs.






\section{Method}

In the main part of our paper, we first briefly discuss the background regarding partial differential equations with a focus on the Navier-Stokes equations and the damped wave equation, before we introduce our physics-informed Hermite spline CNN approach (Spline-PINN), that combines the advantages of PINNs regarding physics-informed training without training data and Spline-CNNs regarding their generalization capability.

\subsection{Partial Differential Equations}\label{ssec:PDE}
Partial Differential Equations describe the dependencies between partial derivatives of a multivariate function inside a domain $\Omega$ and usually need to be solved for given initial conditions at the beginning of the simulation and boundary conditions at the domain boundaries $\partial \Omega$. 

\subsubsection{The Incompressible Navier-Stokes Equations}\label{ssec:INS} 
describe the dynamics of a fluid with a pressure-field $p$ and a velocity-field $\vec{v}$ inside a domain $\Omega$ by means of the incompressibility and momentum equation:

\begin{align}
    \nabla \cdot \vec{v} &= 0 & \textrm{in }\Omega\\
    \rho \dot{\vec{v}} &= \rho (\partial_t \vec{v} + \vec{v} \cdot \nabla \vec{v}) = \mu \Delta \vec{v} - \nabla p + \vec{f} &\textrm{in } \Omega
\end{align}
Here, $\rho$ represents the fluid's density and $\mu$ its viscosity. The external forces $\vec{f}$ are neglected in our experiments. These two equations have to be solved for given initial conditions $\vec{v}_0,p_0$ and boundary conditions (BCs). Here, we consider the Dirichlet BC to set the velocity field $\vec{v} = \vec{v}_d$ at the domain boundaries $\partial \Omega$. 
In this work, we exploit the Helmholtz decomposition theorem and use a vector potential $\vec{a}$ with $\vec{v} = \nabla \times \vec{a}$ to automatically ensure incompressibility. 
Furthermore, we restrict our considerations to 2D flows and, hence, only the $z$-component of $\vec{a}$, $a_z$, is needed. %
The Navier-Stokes equations are considered particularly hard to solve due to the non-linear advection term ($\vec{v}\cdot \nabla \vec{v}$) and, therefore, are the main focus of our investigations.  

\subsubsection{The Damped Wave Equation} \label{ssec:DW}
can be used to describe for example the dynamics of a thin membrane with height-field $z$ and velocity-field $v_z$.
\begin{align}
    \partial_t z &= v_z &\textrm{in }\Omega \\
    \partial_t v_z &= k \Delta z - \delta v_z &\textrm{in }\Omega
\end{align}
Here, $k$ is the stiffness constant of the membrane and $\delta$ is a damping constant. As for the Navier-Stokes equations, we solve these equations for given initial conditions $z_0,v_{z0}$ and Dirichlet boundary conditions: $z = z_d \textrm{ on }\partial \Omega$. 

Although the damped wave equation contains only linear components, we present this additional example to demonstrate that our method might also work for different classes of PDEs. The wave equation lays the foundation for many more complex equations such as the electromagnetic wave equation or the Schrödinger equation. Furthermore, for high damping constants $\delta$, $z$ converges towards the solution of the Laplace equation. 

\subsection{Spline-PINN}
In this section, we introduce our Physics-Informed Hermite Spline CNN. First, we provide an overview on Hermite splines and then show how to incorporate them into a CNN for solving PDEs.

\subsubsection{Hermite Splines}

\begin{figure*}[ht]
\centering
\includegraphics[width=1.8\columnwidth]{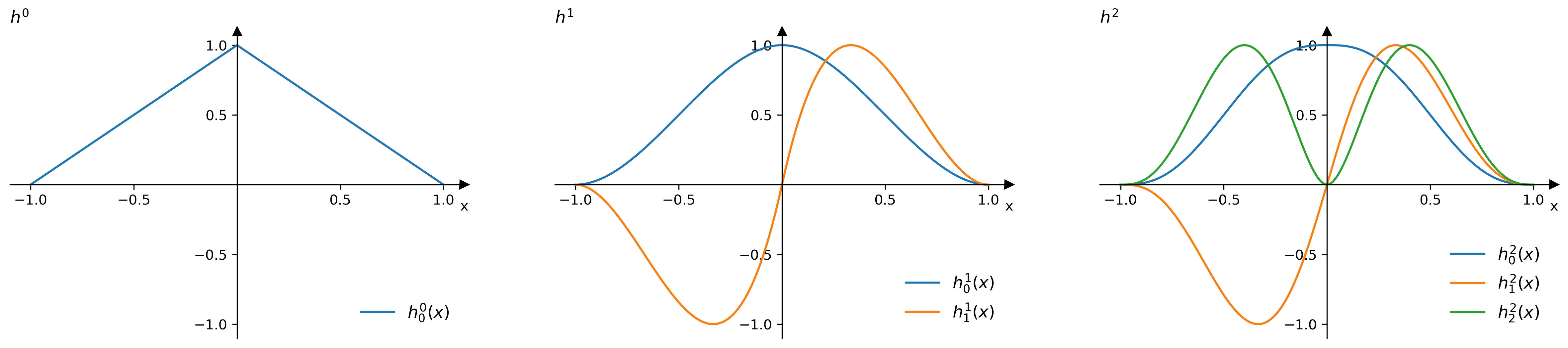} 
\caption{1D Hermite spline kernels for $n=0,1,2$ (scaled between -1 and 1). Note, that these kernel functions are in $C^n$ and thus, the $(n+1)$-th derivatives are of bounded variation. 
The case $n=0$ could be considered as a linear interpolation between spline coefficients.}
\label{hermite_spline_1D}
\end{figure*}

In general, Hermite splines are piecewise polynomials that are defined by their values and their first $n$ derivatives at given support points. To facilitate CNN-based processing, we arrange the support points on a regular grid. Figure \ref{hermite_spline_1D} shows an example of Hermite spline kernel functions in 1D for $n=0,1,2$. We define the kernel functions $h_i^n(x)$, such that the values and first $n$ derivatives at the support points ($x=-1,0,1$) are set to 0 except the $i$th derivative at $x=0$ which takes a value such that $h_i^n(x)\in[-1,1]$. 
In contrast to B-Spline kernels, the support of Hermite spline kernels ranges only over the directly neighboring grid cells. 
This facilitates learning of high frequencies and allows for computationally more efficient interpolations with comparatively small transposed convolution kernels. 
By linearly combining Hermite spline kernel functions of order $n$ at every grid-cell we obtain continuous piecewise polynomials with $(n+1)$-th derivatives of bounded variation. Choosing the right spline order $n$ is important to be able to compute the physics informed loss for a given PDE as will be discussed below.

To obtain kernel functions in multiple dimensions, we use the tensor product of multiple 1D kernel functions:
\begin{equation}
    h^{l,m,n}_{i,j,k}(x,y,t) = h^l_i(x)h^m_j(y)h^n_k(t)
\end{equation}
In the supplementary, we show how basis flow fields can be obtained by taking the curl of $h^{l,m}_{i,j}(x,y)$. 
On a grid $\hat x, \hat y,\hat t \in \hat X\times \hat Y \times \hat T$ with discrete spline coefficients $c^{i,j,k}_{\hat x,\hat y,\hat t}$, we obtain a continuous Hermite spline $g(x,y,t)$ as follows:
\begin{equation}
    g(x,y,t) = \sum_{\substack{i,j,k \in [0:l]\times[0:m]\times[0:n]\\\hat x, \hat y,\hat t \in \hat X\times \hat Y \times \hat T}} c^{i,j,k}_{\hat x,\hat y,\hat t} h^{l,m,n}_{i,j,k}(x-\hat x,y-\hat y,t-\hat t) \label{eq:spline_interpolation}
\end{equation}
Our goal is, to find spline coefficients $c^{i,j,k}_{\hat x,\hat y,\hat t}$ such that $g(x,y,t)$ matches the solution of PDEs as closely as possible. The partial derivatives of $g$ with respect to $x,y,t$ can be directly obtained by taking the corresponding derivatives of the spline kernel functions. 

\subsubsection{Pipeline}
\begin{figure}[h]
\centering
\includegraphics[width=0.9\columnwidth]{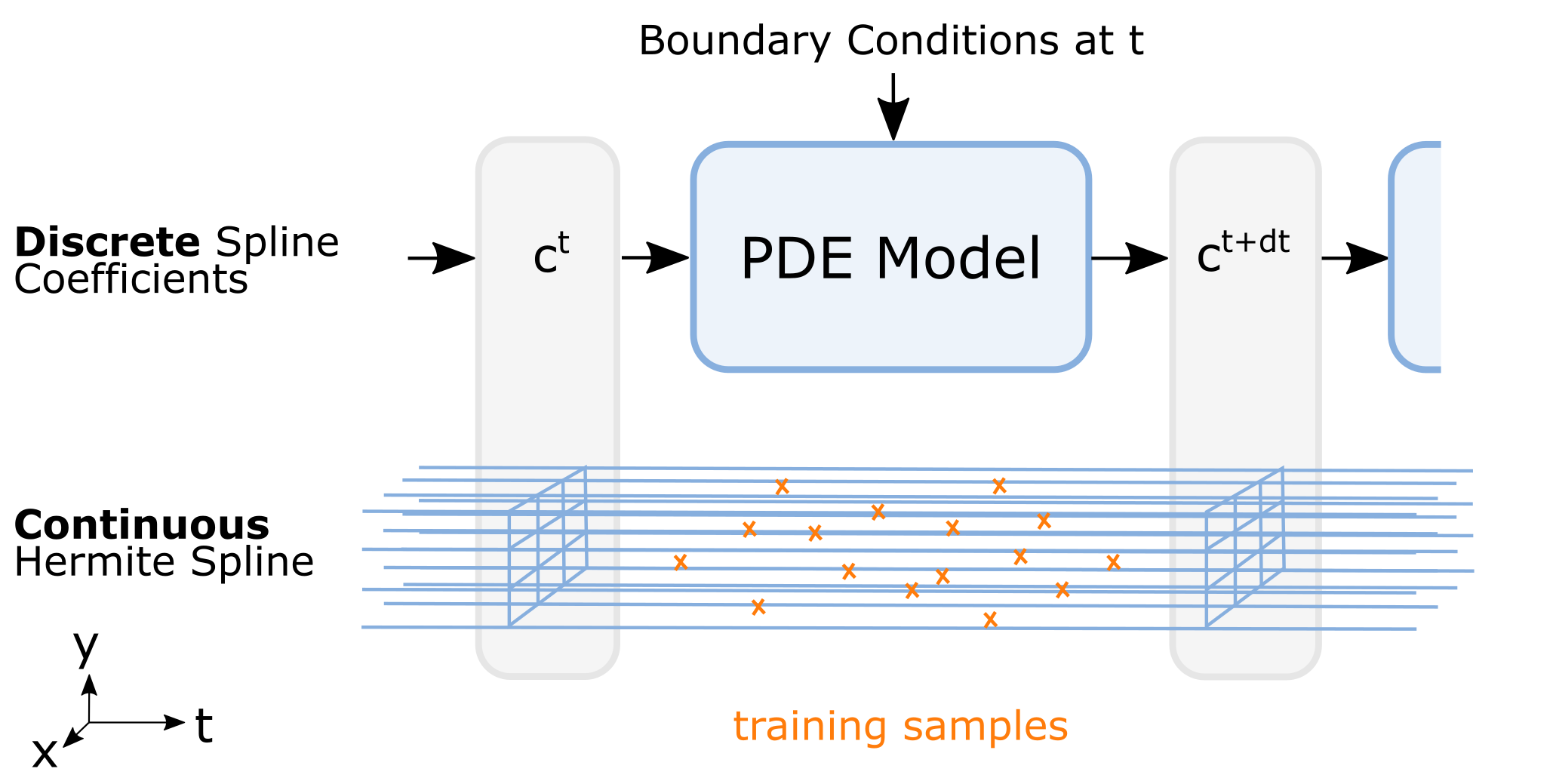}
\caption{Pipeline of PDE Model with Hermite spline interpolation. Since the solution is continuous, training samples and evaluation samples can be obtained at any point in space and time. More detailed views of the PDE Models used for the Navier-Stokes equations and damped wave equation are provided in the supplementary material.}
\label{pipeline}
\end{figure}

Figure \ref{pipeline} depicts the pipeline of our Spline-PINN. We use a CNN (PDE Model) to map discrete Hermite-spline coefficients and boundary conditions from a timepoint $\hat t$ to spline coefficients at a succeeding timepoint $\hat t + dt$. By recurrently applying the PDE Model on the spline coefficients, the simulation can be unrolled in time. From these discrete spline-coefficients, continuous Hermite splines can be efficiently obtained using transposed convolutions (see Equation \ref{eq:spline_interpolation}) with interpolation kernels as depicted in Figure~\ref{hermite_spline_1D}. This way, training samples as well as evaluation samples can be taken at any point in space and time. 



\subsubsection{Physics Informed Loss}
\begin{figure*}[h]
\centering
\includegraphics[width=1\linewidth]{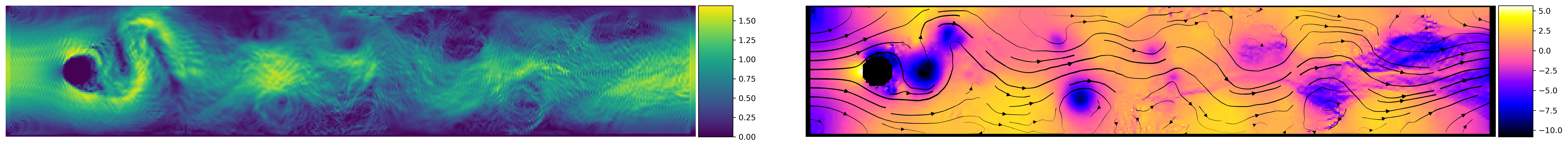}
\caption{Severe discretization artifacts appear using a physics-constrained loss based on a finite differences Marker and Cell grid at $Re=10000$ - even at twice the resolution compared to our Hermite spline approach.} 
\label{MAC_artifacts}
\end{figure*}

When training neural networks on PDEs, the information provided by the PDEs themselves can be used to save or even completely spare out training-data. 
In literature, two major approaches have been established: 

Physics-\emph{constrained} approaches \cite{Zhu:2019} compute losses \emph{explicitly} - usually on a grid topology using finite differences in order to approximate the derivatives of a PDE. This approach is suitable to train CNNs and allows neural networks to learn e.g. the incompressible Navier-Stokes equations without any ground truth training-data and generalize to new domains~\cite{Wandel:2020}. 
However, relying on finite differences may lead to inaccuracies and strong discretization artifacts - especially at high Reynolds-numbers (see Figure~\ref{MAC_artifacts}). 

In contrast, physics-\emph{informed} approaches \cite{Raissi:2019} compute losses \emph{implicitly} - usually by taking derivatives of an implicit field description. 
This approach enables efficient training of implicit neural networks and yields continuous solutions that can be evaluated at any point in space and time. However, implicit neural networks do not generalize well to novel domains but usually require retraining of the network.

Here, we can combine the advantages of both approaches: By using a convolutional neural network that processes spline coefficients which can be considered as a discrete hidden latent description for a continuous implicit field description based on Hermite splines, our Spline-PINN approach is capable to generalize to new domain geometries and yields continuous solutions that avoid the detrimental effects of a discretized loss function based on finite differences. 

We aim to optimize the spline coefficients such that the integrals of the squared residuals of the PDEs over the domain / domain-boundaries and time steps are minimized. To compute these integrals, we uniform randomly sample points within the given integration domains. 

For the Navier-Stokes equation, we consider a momentum loss term $L_p$ defined as:
\begin{equation}
    L_p = \int_\Omega \int_{\hat{t}}^{\hat{t}+dt} ||\rho (\partial_t \vec{v} + \vec{v} \cdot \nabla \vec{v}) - \mu \Delta \vec{v} + \nabla p||^2
\end{equation}
It is important that the residuals are of bounded variation, otherwise we can not compute the integral. $L_p$ contains third order derivatives for $a_z$ in space (see viscosity term: $\mu \Delta (\nabla \times a_z)$) and thus, following the argument in Figure \ref{hermite_spline_1D}, we have to choose at least $l,m=2$ for the Hermite spline kernels of $a_z$.
The boundary loss term $L_b$ is given by:
\begin{equation}
    L_b = \int_{\partial \Omega} \int_{\hat{t}}^{\hat{t}+dt} ||\vec{v}-\vec{v}_d||^2
\end{equation}
$L_p$ and $L_b$ are then combined in the final loss term $L_\textrm{tot}^\textrm{flow}$ according to:
\begin{equation}
    L_\textrm{tot}^\textrm{flow} = \alpha L_p + \beta L_b
\end{equation}
Here, we set hyperparameters $\alpha=10$ and $\beta=20$. $\beta$ was chosen higher compared to $\alpha$ to prevent the flow field from leaking through boundaries. A more thorough discussion regarding the choice of hyperparameters is provided in the supplementary material.

For the damped wave equation, we need two loss terms inside the domain $\Omega$:
\begin{align}
    L_z &= \int_\Omega \int_{\hat{t}}^{\hat{t}+dt} ||\partial_t z-v_z||^2\\
    L_v &= \int_\Omega \int_{\hat{t}}^{\hat{t}+dt} ||\partial_t v_z - k\Delta z + \delta v_z||^2
\end{align}
Similar to the incompressible Navier-Stokes equation, we have second order derivatives of $z$ in space (see $k\Delta z$-Term) and thus have to choose at least $l,m=1$ for the Hermite spline kernels of $z$.
The boundary loss term is:
\begin{equation}
    L_b = \int_{\partial \Omega} \int_{\hat{t}}^{\hat{t}+dt} ||z-z_d||^2
\end{equation}
$L_z,L_v$ and $L_b$ are then combined in the final loss term $L_\textrm{tot}^\textrm{wave}$:
\begin{equation}
    L_\textrm{tot}^\textrm{wave} = \alpha L_z + \beta L_v + \gamma L_b
\end{equation}
For the wave equation, we set the hyperparameters $\alpha=1, \beta=0.1, \gamma=10$.

\subsubsection{Training Procedure}
\begin{figure}[h]
\centering
\includegraphics[width=0.9\columnwidth]{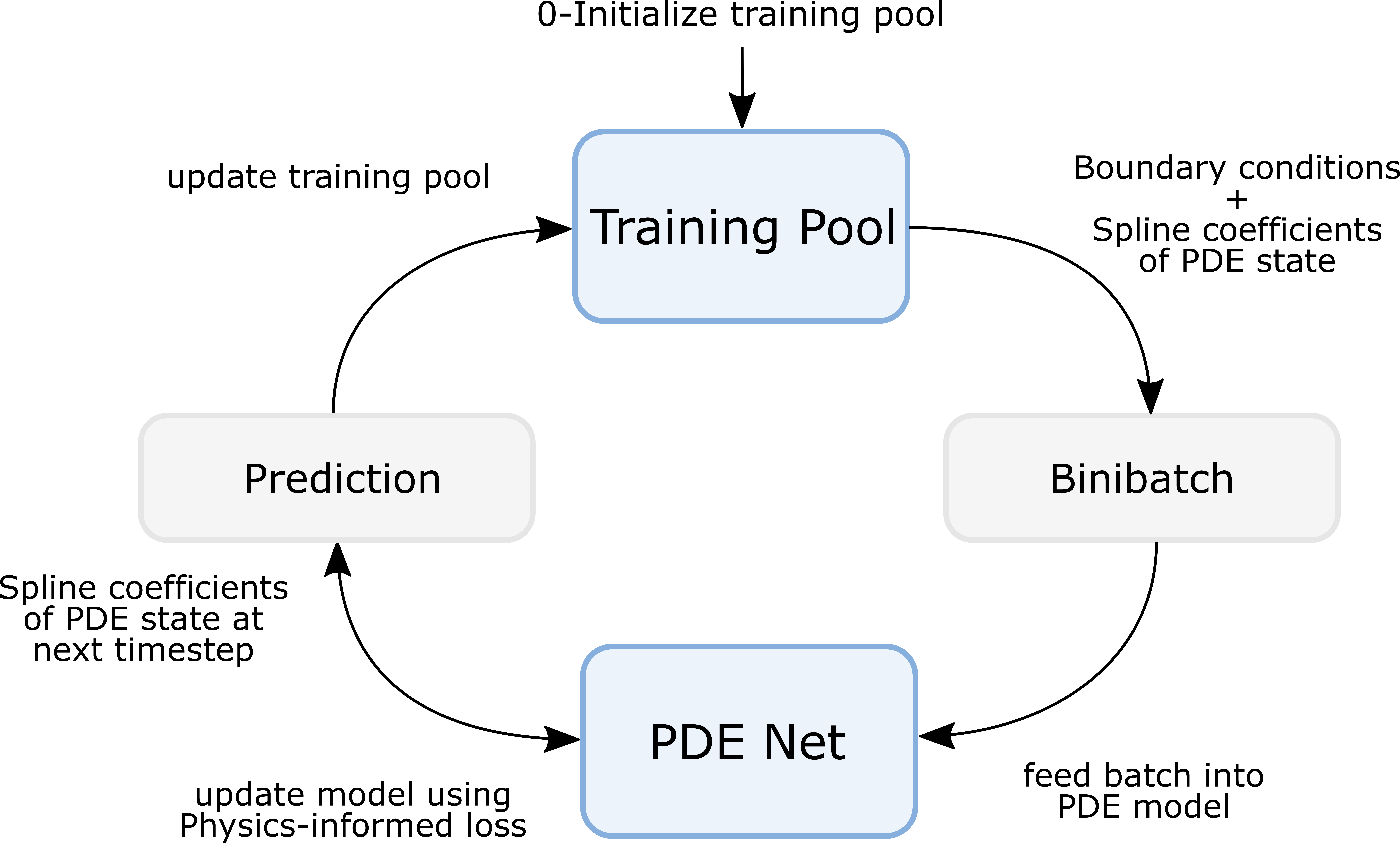}
\caption{Training cycle similar to \cite{Wandel:2020}} 
\label{fig:training_cycle}
\end{figure}

As presented in Figure~\ref{fig:training_cycle}, we use a training procedure similar to \cite{Wandel:2020} but replace the physics-constrained loss based on finite differences by the just introduced physics-informed loss on the spline coefficients. 
First, we initialize a training pool of randomized training domains and spline coefficients. At the beginning, all spline coefficients can be set to 0, thus no ground truth data is required. Then, we draw a random minibatch (batchsize = 50) from the training pool and feed it into the PDE Model, which predicts the spline coefficients for the next time step. Then, we compute a physics-informed loss inside the volume spanned by the spline coefficients of the minibatch and the predicted spline coefficients in order to optimize the weights of the PDE Model with gradient descent. To this end, we use the Adam optimizer (learning rate = 0.0001). Finally, we update the training pool with the just predicted spline coefficients in order to fill the pool with more and more realistic training data over the course of training. From time to time, we reset spline coefficients to 0 in the training pool in order to also learn the warm-up phase from 0 spline coefficients. 
Training took 1-2 days on a NVidia GeForce RTX 2080 Ti.

\section{Results}

In the following, we provide qualitative and quantitative evaluations of our approach for fluid and wave simulations. Additional results as well as demonstrations where the user can interact with the PDE by dynamically changing boundary conditions, are contained in our supplemental.

\subsection{Fluid Simulation}
\subsubsection{Qualitative Evaluation}
The dynamics of a fluid depends strongly on the Reynolds number, which is a dimensionless quantity that relates the fluid density $\rho$, mean velocity $||\vec{v}||$, obstacle diameter $L$ and viscosity $\mu$ in the following way: 

\begin{equation}
    Re = \frac{\rho ||\vec{v}||L}{\mu}
\end{equation}

To validate our method, we reconstructed the DFG benchmark setup~\cite{cfd-benchmark} and scaled its size and viscosity by a factor of 100. This way, we obtained a grid size of 41 $\times$ 220 cells for the fluid domain and the obstacle diameter is 10 cells. 

\begin{figure*}[htb]
  \centering
  \includegraphics[width=1\linewidth]{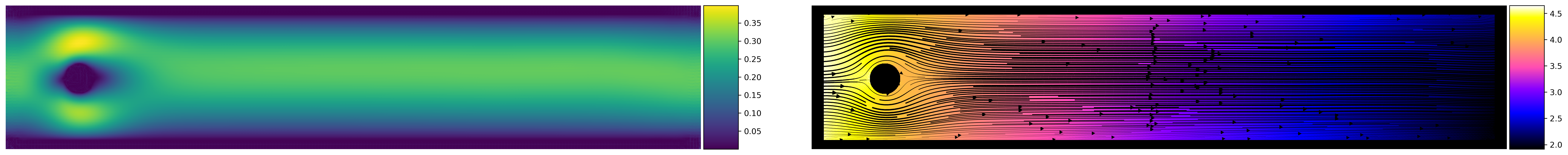}
  \parbox[t]{.9\columnwidth}{(a) time-reversible flow($Re=2, \mu=1, \rho=1$)\label{fig:Re_2}}
  \includegraphics[width=1\linewidth]{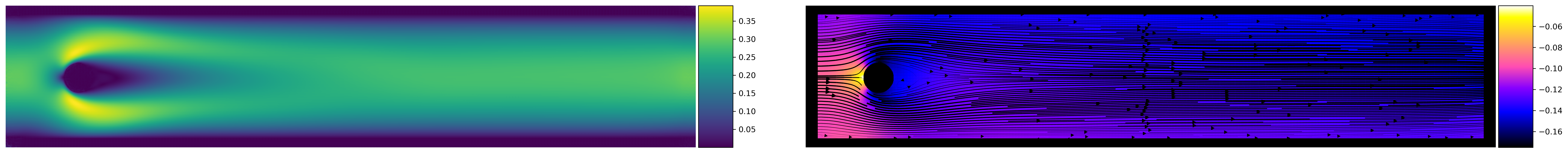}
  \parbox[t]{.9\columnwidth}{(b) laminar flow ($Re=20, \mu=0.1, \rho=1$)\label{fig:Re_20}}
  \includegraphics[width=1\linewidth]{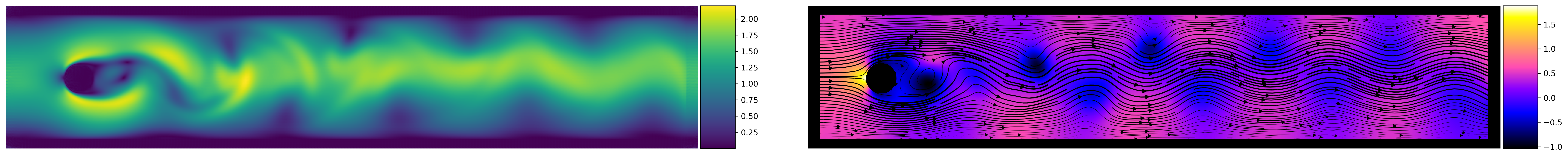}
  \parbox[t]{.9\columnwidth}{(c) laminar vortex street($Re=100, \mu=0.1, \rho=1$)\label{fig:Re_100}}
  \includegraphics[width=1\linewidth]{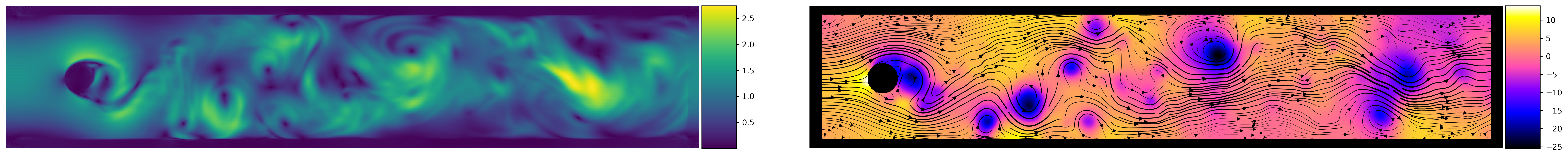}
  \parbox[t]{.9\columnwidth}{(d) turbulent flow($Re=1000, \mu=0.1, \rho=10$)\label{fig:Re_1000}}
  \includegraphics[width=1\linewidth]{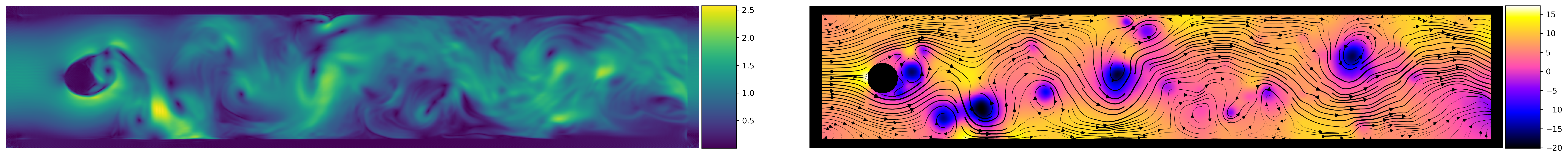}
  \parbox[t]{.9\columnwidth}{(e) turbulent flow ($Re=10000, \mu=0.01, \rho=10$)\label{fig:Re_10000}}
  \caption{\label{fig:cylinder_1}
           Flow and pressure fields around a cylinder obtained by our method at different Reynolds numbers. Left side: velocity magnitude; Right side: pressure field and stream lines of velocity field. An animated real-time visualization of these experiments is provided in the supplementary video (\url{https://www.youtube.com/watch?v=QC98LCtCZn0}). }
\end{figure*}

For very small Reynolds numbers, the flow field becomes basically time-reversible which can be recognized by the symmetry of the stream lines in Figure~\ref{fig:cylinder_1} a). At $Re=20$ a steady wake is forming behind the obstacle (see Figure~\ref{fig:cylinder_1} b). For higher Reynolds numbers such as $Re=100$, this wake becomes unstable and a von Karman vortex street is forming (see Figure~\ref{fig:cylinder_1} c). Figure~\ref{fig:cylinder_1} d) and e) show turbulent flow fields at very high Reynolds numbers ($Re=1000$ and $Re=10000$). Notice the high level of details in the velocity field learned by our method without any ground truth data. 

\subsubsection{Quantitative Evaluation}

\begin{figure}[h]
\centering
\includegraphics[width=1\columnwidth]{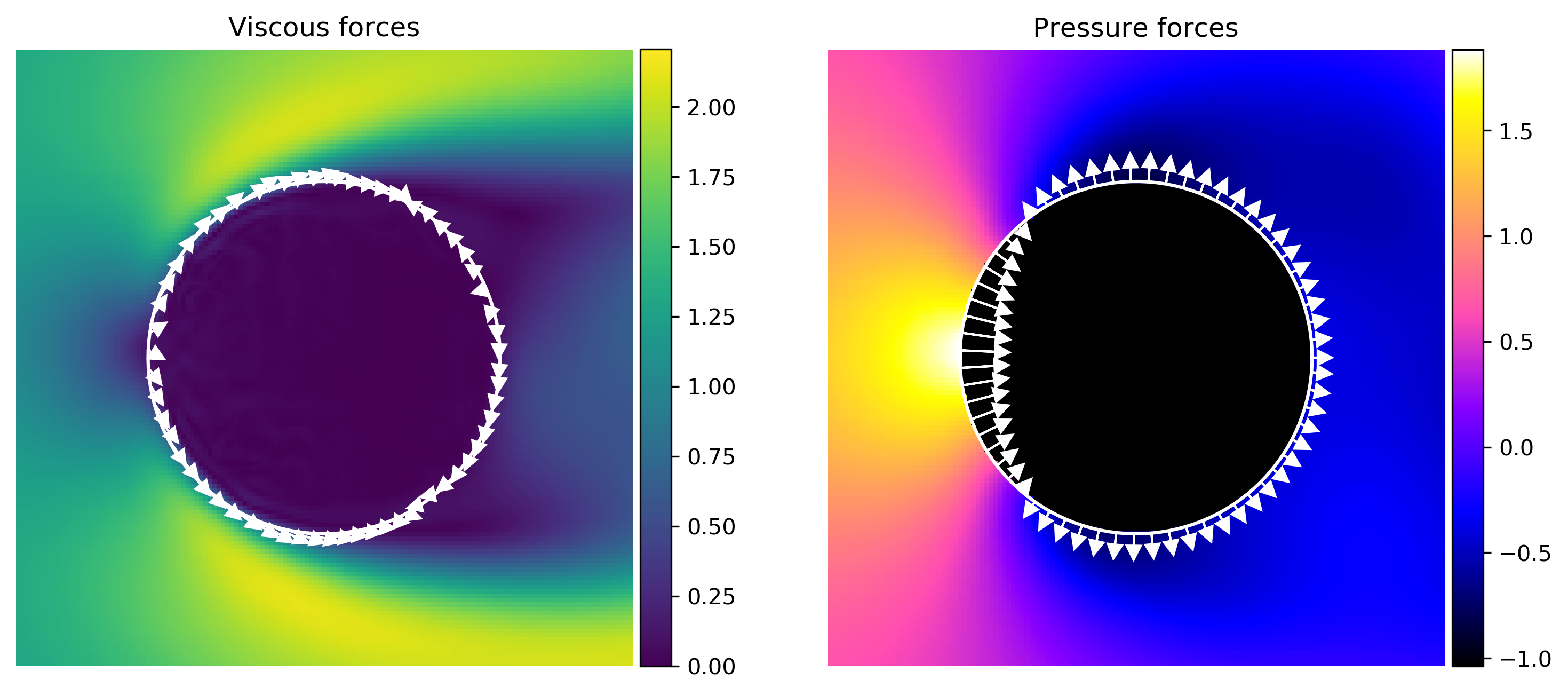}
\caption{White arrows indicate viscous and pressure forces (see Equation \ref{force_integral}) acting on a cylinder at $Re=100$ (see Figure \ref{fig:cylinder_1} c). While forces from viscous friction are parallel to the obstacle's surface, pressure forces are always perpendicular to the surface.} 
\label{fig:forces}
\end{figure}

To evaluate our method quantitatively, we computed the forces exerted by the fluid on a cylinder (see Figure \ref{fig:forces}) and compared the resulting drag and lift coefficients to an implicit physics-informed neural network similar to \cite{raissi2018hidden}, 
a MAC grid based physics-constrained neural network \cite{Wandel:2020}, an industrial CFD solver (Ansys) and official results reported on a CFD benchmark ~\cite{cfd-benchmark}. To the best of our knowledge, this is the first time such forces were obtained by a fully unsupervised machine learning based approach. The forces can be separated into two terms stemming from  viscous friction ($\vec{F}_\mu$) and  the pressure field ($\vec{F}_p$):
\begin{align}
    \vec{F}_\mu &= \int_S \mu (\nabla \vec{v}) \vec{n} ds \, ; \, \vec{F}_p = \int_S -p \vec{n} ds\label{force_integral}\\
    \vec{F}_{\textrm{tot}} &= \vec{F}_\mu + \vec{F}_p
\end{align}
Figure \ref{fig:forces} shows the distribution of such viscous drag and pressure forces along the cylinder surface $S$ with surface normals $\vec{n}$.

The drag-force $F_D$ is the parallel force-component of $\vec{F}_{\textrm{tot}}$ to the flow direction ($F_D=\vec{F}_{\textrm{tot},x}$) while the lift force $F_L$ is its orthogonal component ($F_L=\vec{F}_{\textrm{tot},y}$). From these forces, it is possible to compute drag and lift coefficients as follows:
\begin{equation}
    C_D = \frac{2 F_D}{U^2_\textrm{mean}L},\,
    C_L = \frac{2 F_L}{U^2_\textrm{mean}L}
\end{equation}

\begin{figure}[h]
\centering
\includegraphics[width=1\columnwidth]{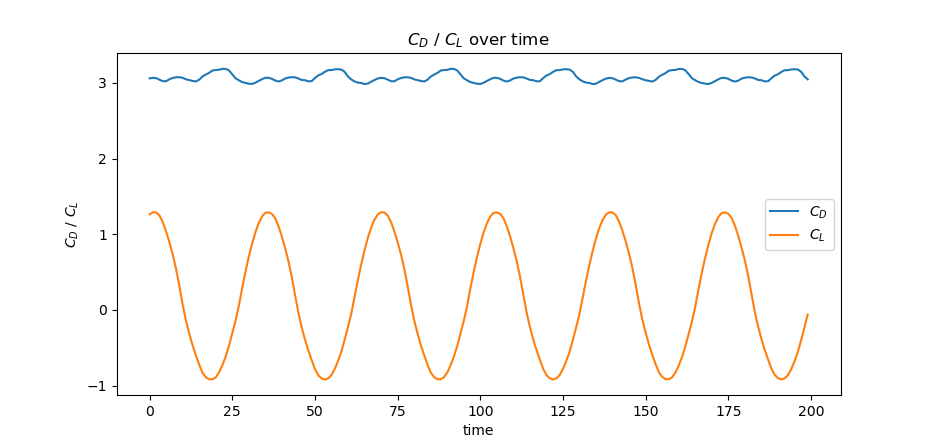}
\caption{Oscillating drag and lift coefficients over time obtained by our Spline-PINN at $Re=100$.}
\label{fig:cd_cl_over_time}
\end{figure}

Table \ref{tab:cd_cl} compares the drag and lift coefficients obtained by our method to an implicit PINN, 
a finite-difference based MAC grid approach~\cite{Wandel:2020}, an industrial CFD solver (Ansys) and the official DFG benchmark~\cite{cfd-benchmark} for Reynolds numbers 2, 20 and 100. For all Reynolds numbers, the Spline-PINN approach returned significantly improved drag coefficients compared to the implicit PINN and MAC-approach. For $Re=100$, the implicit PINN fails to capture the dynamics of the von Karman vortex street as it is only trained on the domain boundaries without any additional information stemming for example from a moving die \cite{raissi2018hidden} or dynamic boundary conditions \cite{Jin_2021}. In contrast, our method is able to reproduce such oscillations as can be seen in Figure \ref{fig:cd_cl_over_time}. The implicit PINN required retraining for every setting which took about 1 day while performing the computations with the MAC grid based approach or our Spline-PINN was achieved basically in real-time at 60 and 30 iterations per second respectively. The Ansys solver took 9 sec for $Re=2$, 13 sec for $Re=20$ and 37 min for $Re=100$. 
A comparison of errors in $E[|\nabla\cdot \vec{v}|]$ and $L_p$ as well as a stability analysis can be found in the supplemental.

\begin{table*}[h]
\centering
\small
\begin{tabular}{c|c|c|c|c|c|c|c|}
    & \multicolumn{2}{c|}{Re=2}& \multicolumn{2}{c|}{Re=20}& \multicolumn{3}{c|}{Re=100}\\
    Method & $C_D$ & $C_L$ & $C_D$ & $C_L$ & $C_D$ & $C_L$ & \textrm{time} \\
     \hline
    implicit PINN & 25.3 & 0.478 & 3.299& 0.0744 & 1.853 & -0.02445 & $\sim$ 1 day \\ 
    MAC grid~\cite{Wandel:2020} & 25.76 & -0.824 & 4.414 & -0.597 & (2.655 / 2.693 / 2.725)* & (-0.757 / 0.0184 / 0.86)* & $\sim$ 15 sec\\
    Spline-PINN (ours) & 29.7 & -0.456 & 4.7 & 5.64e-04 & (2.985 / 3.068 / 3.188)* & (-0.926 / 0.179 / 1.295)* & $\sim$ 10 sec\\
    Ansys & 32.035 & 0.774 & 5.57020 & 0.00786 & (3.234 / 3.273 / 3.31)* & (-1.14 / -0.059 / 1.07)* & 37 min \\
    DFG-Benchmark~\cite{cfd-benchmark} & - & - & 5.58 & 0.0106 & (3.1569 / 3.1884 / 3.220 )*& (-1.0206 / -0.0173 / 0.9859)* & -
\end{tabular}
\caption{Drag and lift coefficients obtained by an implicit PINN, our Hermite spline approach, an industrial CFD solver (Ansys) and official results from the DFG-benchmark ~\cite{cfd-benchmark}. *:(minimum/average/maximum)-values for oscillating coefficients}
\label{tab:cd_cl}
\end{table*}

\subsection{Wave Simulation}
\subsubsection{Qualitative Evaluation}
\begin{figure}[h!]
  \centering
  \includegraphics[width=0.87\linewidth]{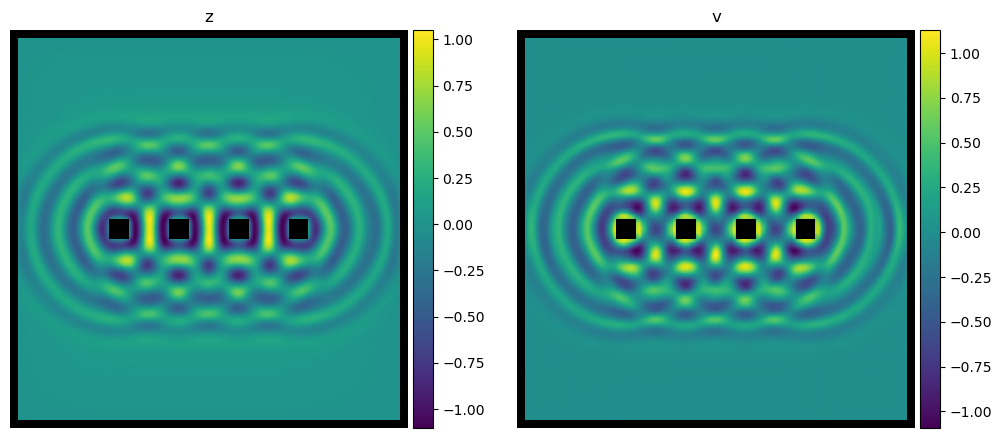}
  \parbox[t]{.9\columnwidth}{(a) Interference patterns forming around 4 oscillators.\label{fig:interference}}
  \includegraphics[width=0.87\linewidth]{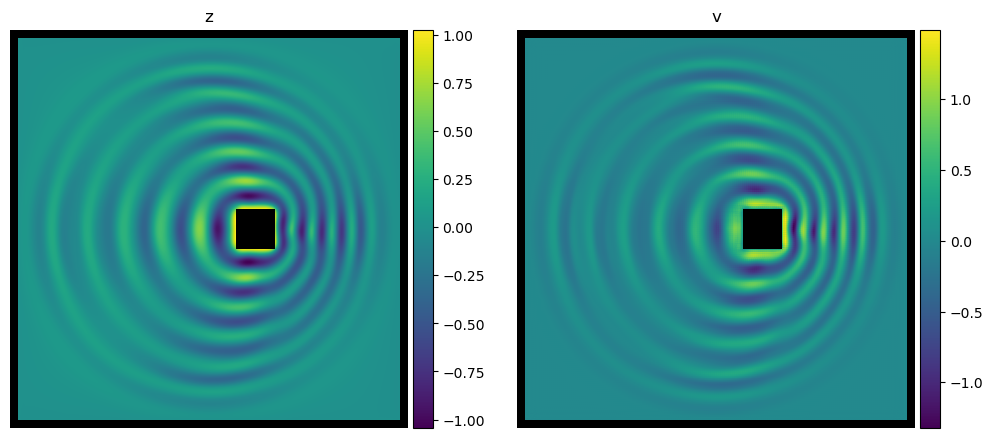}
  \parbox[t]{.9\columnwidth}{(b) Doppler effect of an oscillator moving to the right.\label{fig:doppler}}
  \includegraphics[width=0.87\linewidth]{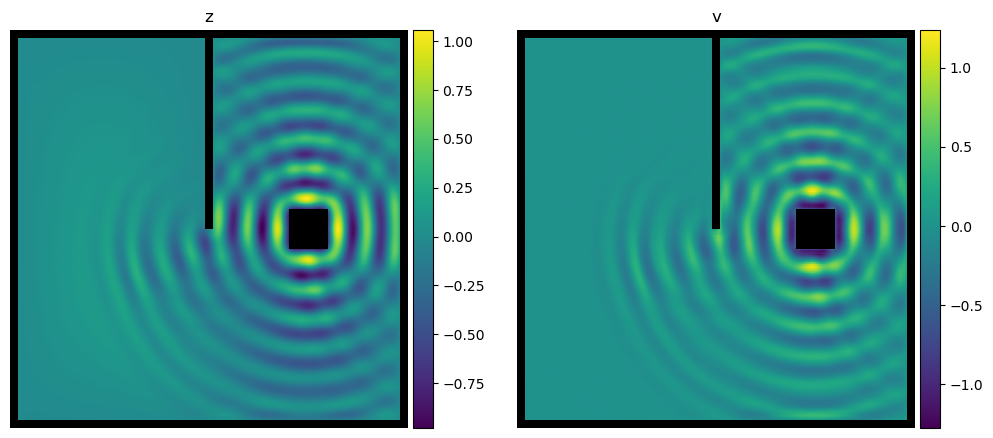}
  \parbox[t]{.9\columnwidth}{(c) Wave reflections on domain boundaries.\label{fig:reflection}}
  \caption{\label{fig:waves}
           Results of our method for the wave equation ($k=10,\delta=0.1$). Left side: height field $z$; Right side: velocity field $v_z$. The domain boundaries are marked in black. For better performance, second-order splines were used in $x$ and $y$. An animated real-time visualization of these experiments can be found at \url{https://www.youtube.com/watch?v=QC98LCtCZn0}.}
\end{figure}

Figure~\ref{fig:waves} shows several experimental results that were obtained by our method for the damped wave equation on a 200 $\times$ 200 domain. First, we investigated interference patterns that arise when waves from different directions are superimposed. In Figure \ref{fig:waves} a), the waves of 4 oscillators interfere with each other. Then, in Figure \ref{fig:waves} b), we investigate whether our method is able to learn the Doppler effect for moving oscillators. This effect is well-known e.g. for changing the pitch of an ambulance-car that drives by. Finally, in Figure \ref{fig:waves} c), we show the reflection and interference behavior of waves hitting Dirichlet boundaries. 

As for the fluid simulations, all of these results were obtained without relying on any ground truth data. Furthermore, the domain in Figure \ref{fig:waves} c) was not contained in the randomized set of training domains indicating good generalization performance. 

\subsubsection{Quantitative Evaluation}
Table \ref{tab:wave_loss_quant} compares the losses of our method on the oscillator domain (see Figure \ref{fig:waves} a). We trained two versions with two different spline orders in the spatial dimensions ($l,m=1$ and $l,m=2$) and observed significantly better performance for $l,m=2$. 
The stability of our approach for the wave equation is examined in the supplementary material.

\begin{table}[h]
\centering
\begin{tabular}{c|c|c|c}
    Spline order & $L_z$ & $L_v$ & $L_b$ \\
     \hline
    $l,m=1$ & 8.511e-02 & 1.127e-02 & 1.425e-03 \\
    $l,m=2$ & 5.294e-02 & 6.756e-03 & 1.356e-03 \\
\end{tabular}
\caption{Quantitative results of wave equation.}
\label{tab:wave_loss_quant}
\end{table}



\section{Conclusion}
In this work, we approach the incompressible Navier-Stokes equation and the damped wave equation by training a continuous Hermite Spline CNN using physics-informed loss functions only. While finite-difference based methods break down at high Reynolds-numbers due to discretization artifacts, our method still returns visually appealing results at $Re=10000$. Furthermore, we investigated drag and lift coefficients on a CFD benchmark setup and observed reasonable accordance with officially reported values which is remarkable given the fact that our method does not rely on any ground truth data.

In the future, further boundary conditions (e.g. von Neumann BC) could be incorporated into the PDE model as well. To further refine the solutions at the boundary layers, a multigrid Hermite spline CNN could be considered. 
The fully differentiable nature of our method may also help in reinforcement learning scenarios, optimal control, sensitivity analysis or gradient based shape optimization.

We believe that our method could have applications in physics engines of computer games or in computer-generated imagery as it provides fast and visually plausible solutions. The obtained drag and lift coefficients indicate that in the future, unsupervised ML based methods could reach levels of accuracy that are sufficiently close to traditional industrial CFD solvers making them suitable for fast prototyping in engineering applications. We firmly believe that moving from explicit physics-constrained losses to implicit physics-informed losses on continuous fields based on discrete latent descriptions such as spline coefficients will positively influence the performance of future ML based PDE solvers that generalize.



\section{Supplementary}

\subsection{Magnus effect and generalization examples}
Figure \ref{fig:magnus_and_generalization} a) demonstrates that our Spline-PINN approach is capable of handling the Magnus effect. The Magnus effect appears e.g. on rotating bodies within a stream and leads to a characteristic low pressure field on the side that moves in the same direction as the flow field. This effect is often exploited in ballsports such as soccer (banana cick) or tennis (topspin) to deflect the path of a ball. Figure \ref{fig:magnus_and_generalization} b) shows a generalization example of the Spline-PINN. During training, we confronted the Spline-PINN with randomized environments that contained at most one obstacle. Nevertheless, our approach is able to generalize to multiple obstacles as shown here by the smiley face.

\subsection{Video - interactive demo}
In our supplementary video, we present dynamic simulations obtained by our Spline-PINN method. All simulations were done in real time on a NVidia GeForce RTX 2080 Ti. To allow for dynamic interactions, we allow the user to dynamically paint new boundaries into the fluid domain during simulation. This proves the high generalization capabilities of our method to scenarios not considered during training. Furthermore, we show dynamic solutions for the wave equation.

\subsection{Stability of Spline-PINNs}

Here, we evaluate the stability of our approach at the example of the incompressible Navier-Stokes equations and the damped wave equation.

\subsubsection{Incompressible Navier-Stokes equation}

\begin{figure}[h!]
\centering
\includegraphics[width=1\columnwidth]{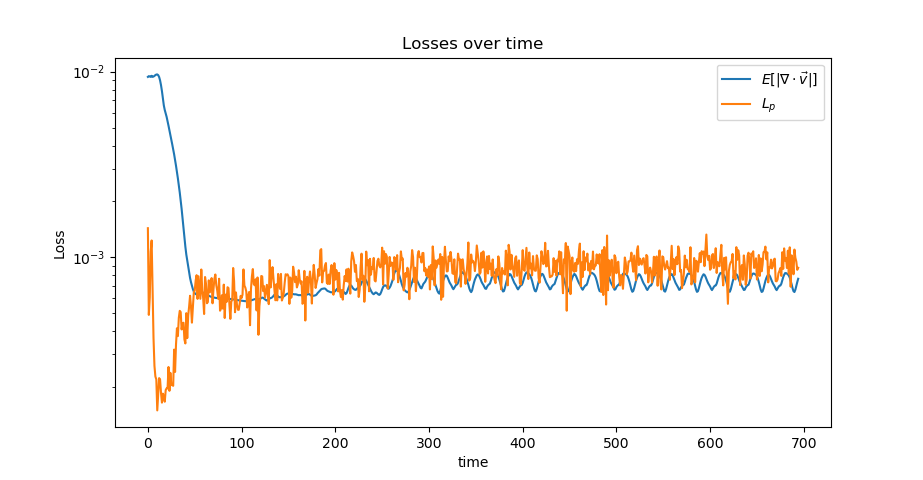}
\caption{Stability of Spline Net while solving the Navier Stokes equation on the DFG benchmark problem at $Re=100$.}
\label{fig:stability_ns_eq}
\end{figure}

We investigated the stability of Spline-PINNs on the incompressible Navier-Stokes equations over hundreds of iterations on the DFG benchmark \cite{cfd-benchmark} problem at $Re=100$. As can be seen in Figure \ref{fig:stability_ns_eq}, after a short warm-up phase of around 50 iterations, in which the network has to start up the simulation from 0 pressure and velocity fields, our method is able to provide stable results over long time horizons.

\subsubsection{Damped wave equation}

\begin{figure}[h!]
\centering
\includegraphics[width=1\columnwidth]{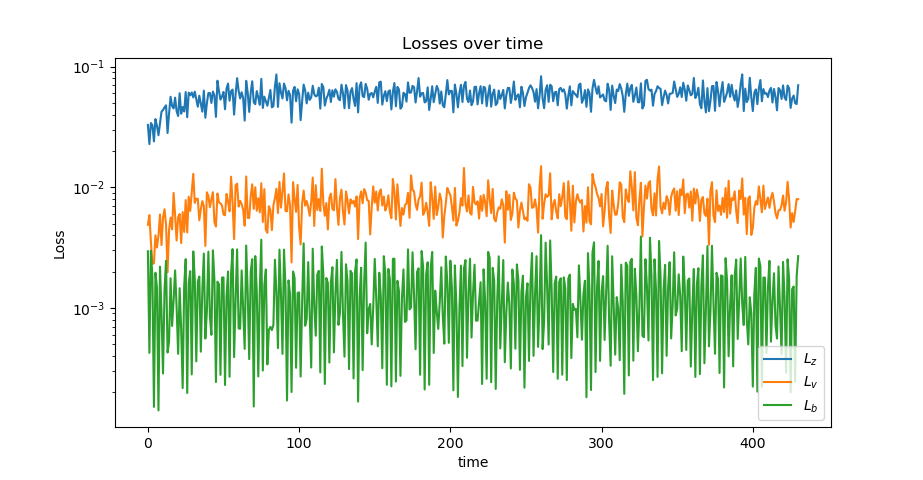}
\caption{Stability of solution for wave equation ($l,m=2$) on the interference problem setup.}
\label{fig:stability_wave}
\end{figure}

\begin{figure*}[h!]
  \centering
  \includegraphics[width=1\linewidth]{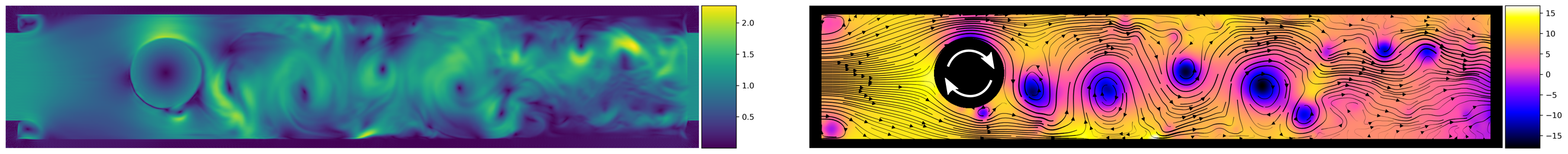}
  \parbox[t]{1\linewidth}{(a) Magnus effect on clockwise turning cylinder ($\mu=0.01, \rho=10$)\label{fig:Magnus}}
  \includegraphics[width=1\linewidth]{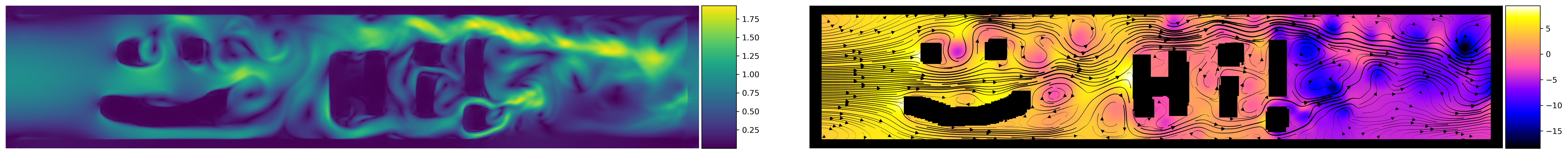}
  \parbox[t]{1\linewidth}{(b) Generalization example (drawing of smiley) ($\mu=0.1, \rho=10$)\label{fig:Smiley}}
  \caption{\label{fig:magnus_and_generalization}
           Flow and pressure fields for Magnus effect and a generalization example. Left side: velocity magnitude; Right side: pressure field and stream lines of velocity field. }
\end{figure*}

As can be seen in Figure \ref{fig:stability_wave}, our method delivers stable results for the damped wave equation on the interference problem setup (see Figure 8 a) in paper). 
In contrast to the loss curves for the Navier Stokes equation (see Figure \ref{fig:stability_ns_eq}), there is no significant warmup phase as 0 spline coefficients already fulfill the boundary conditions and wave equation at the beginning of the simulation very well.

\subsection{Neural network architecture}

\begin{figure*}[h]
\centering
\includegraphics[width=1\linewidth]{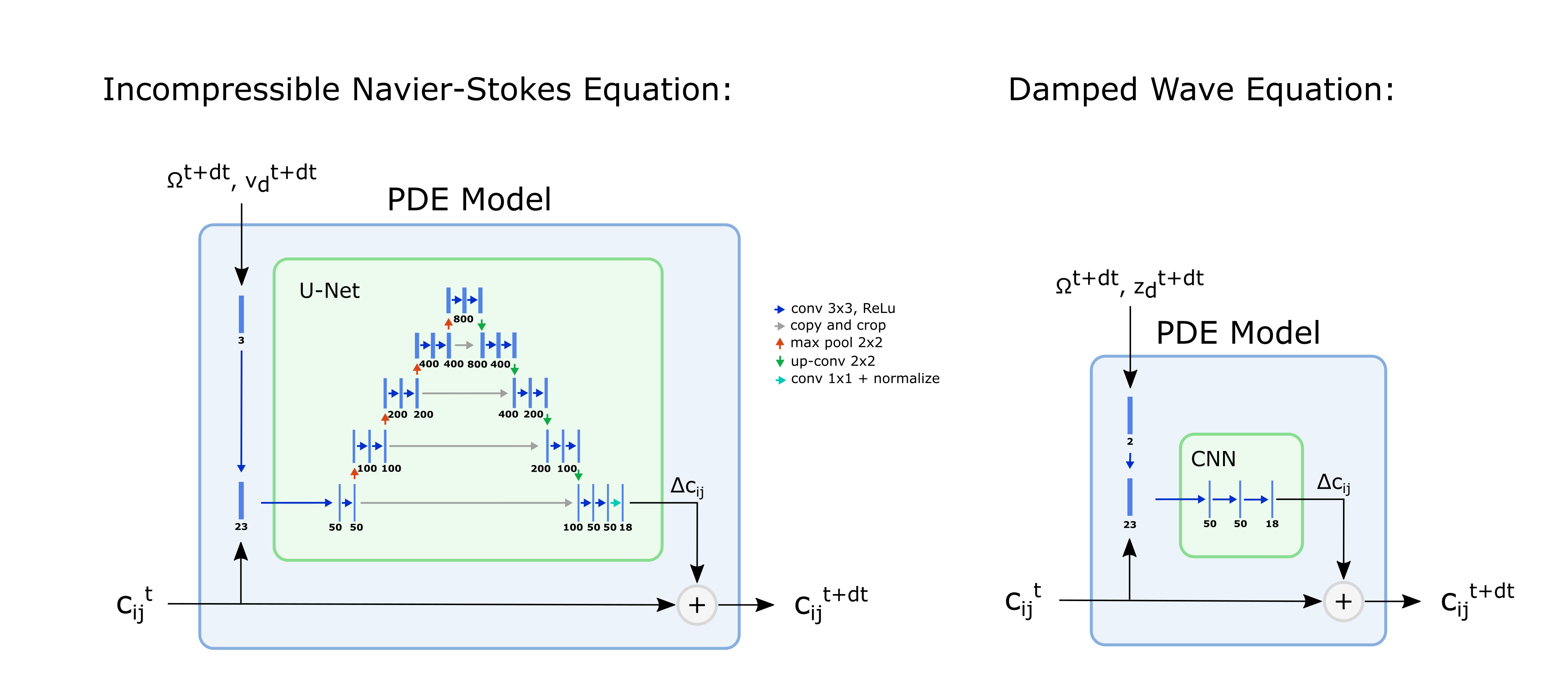}
\caption{Network architectures used for the incompressible Navier-Stokes equation (left) and damped wave equation (right). The numbers located below the blue bars indicate the number of channels of the individual layers.}
\label{fig:net_architecture}
\end{figure*}

In Figure \ref{fig:net_architecture}, we present the network architectures that we used for the incompressible Navier-Stokes equation and the damped wave equation. $\Omega^t$ contains the occupancy grid of the domain at a timepoint $t$ and $\vec{v}_d^t / z_d^t$ contain respective Dirichlet boundary conditions. For the Navier-Stokes equation, we internally rely on a U-Net architecture \cite{ronneberger2015u} while for the damped wave equation, a simple 3-layer CNN was already sufficient. These networks were used to compute a residual ($\Delta c_{i,j}$) to be added to $c_{i,j}^t$ in order to obtain the spline coefficients of the next time-step $c_{i,j}^{t+dt}$. In the case of the Navier-Stokes equations, the vector potential $a_z$ as well as the pressure field $p$ are only defined up to a constant. Thus, we mean-normalize the coefficients of the first Spline-mode for these fields to zero ($\sum_{\hat x, \hat y} c^{0,0,0}_{\hat x,\hat y,\hat t}=0$).

\subsection{Basis flow fields}
\begin{figure}[h]
\centering
\includegraphics[width=0.9\columnwidth]{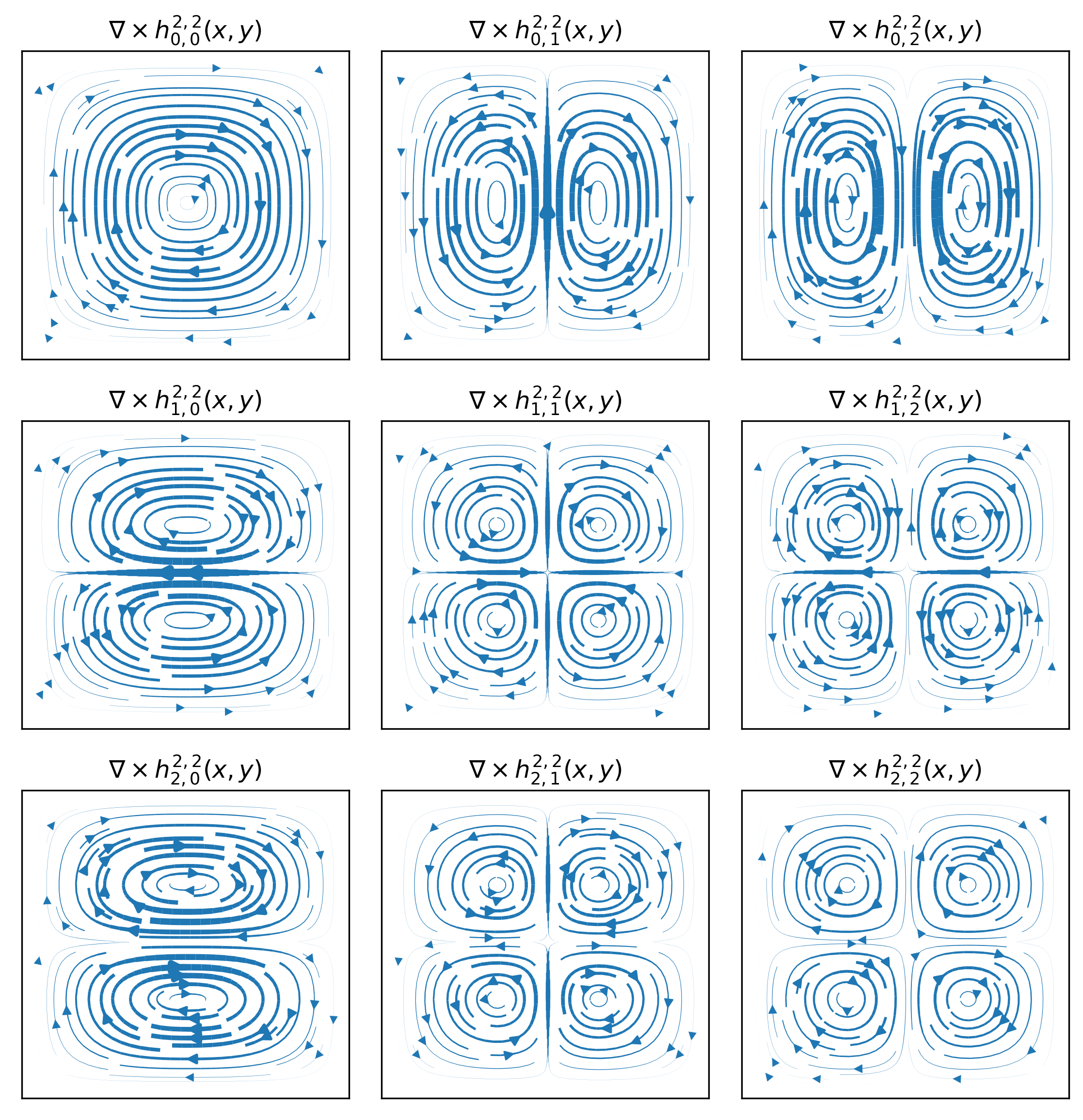}
\caption{Curl of 2D Hermite spline kernels with $l,m=2$. Note that these "basis flow fields" all conserve $\nabla \cdot \vec{v}=0$, since $\nabla \cdot \nabla \times a_z=0$. Thus the total flow field is incompressible as well.}
\label{hermite_spline_2D}
\end{figure}

Figure~\ref{hermite_spline_2D} shows how flow fields can be obtained by taking the curl of individual Hermite spline kernel functions for the vector potential $a_z$. These vector fields could be considered as basis flow fields for the velocity field $\vec{v}$.

\subsection{Choice of hyperparameters $\alpha,\beta,\gamma$}
\begin{figure}[h]
\centering
\includegraphics[width=1\columnwidth]{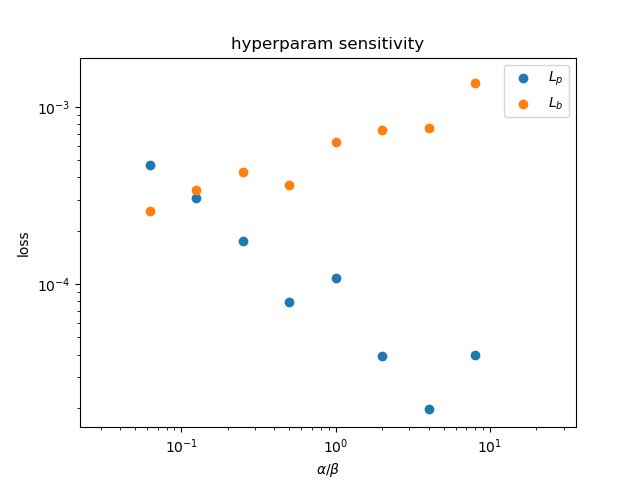}
\caption{Impact of different ratios between $\alpha$ and $\beta$ in $L^\textrm{flow}_\textrm{tot}$ on $L_p$ and $L_b$ after training (see Equation 9).}
\label{fig:hyperparam_sensitivity}
\end{figure}

We investigated the sensitivity of our method with respect to different loss weights in Equation 9. Figure \ref{fig:hyperparam_sensitivity} shows how different ratios between $\alpha$ and $\beta$ affect the losses for the momentum equation $L_p$ and boundary conditions $L_b$ after training. If we put too little weight on the fluid domain ($\alpha$), the accuracy of the momentum equation drops, while if we put too little weight on the boundaries ($\beta$), we get unrealistic flows leaking through the boundary. Tuning these parameters is fairly intuitive and depending on the application, one can choose a corresponding sweet-spot. For the wave equation, one can proceed similarly. Too little weight on $\alpha$ and $\beta$ decreases the performance of the Spline CNN within $\Omega$, while too little weight on the boundaries ($\gamma$) leads to inferior adherence to the boundary conditions. 
In the future, more elaborate methods such as proposed e.g. by Wang et al. \cite{wang2020understanding} could be used to automate the process of finding proper loss weights.

\subsection{Loss values for Navier-Stokes equation}
We compared $E[|\nabla\cdot \vec{v}|]$ and $L_p$ of an implicit physics-informed neural network, a MAC grid based physics-constrained neural network \cite{Wandel:2020}, our Hermite Spline approach and an industrial CFD solver (Ansys) on the DFG Fluid-Benchmark at $Re=2, Re=20$ and $Re=100$. 
All machine learning based approaches were trained without any ground truth data but only using a physics-based loss and the boundary conditions of the benchmark setup \cite{cfd-benchmark}. Since a vector potential is used for the ML-based methods, we computed $E[|\nabla\cdot \vec{v}|]$ by measuring the flow leaking through the boundaries as an effect of not perfectly matching the boundary conditions. This corresponds to a thin layer that does not preserve the incompressibility equation. The losses on the MAC grid were computed using finite differences and the loss for Ansys was computed based on the field-values and derivatives provided by Ansys.

\begin{table*}[h]
\centering
\begin{tabular}{c|c|c|c|c|c|c|}
    & \multicolumn{2}{c|}{Re=2}& \multicolumn{2}{c|}{Re=20}& \multicolumn{2}{c|}{Re=100}\\
    Method & $E[|\nabla\cdot \vec{v}|]$ & $L_p$ & $E[|\nabla\cdot \vec{v}|]$ & $L_p$ & $E[|\nabla\cdot \vec{v}|]$ & $L_p$ \\
     \hline
    implicit PINN & 3.692e-04 & 5.543e-08 & 2.371e-04 & 1.142e-07 & 1.855e-03 & 1.465e-06 \\
    MAC grid & 6.039e-05 & 1.986e-05 & 8.99e-06 & 3.436e-07 & 5.796e-05 & 4.358e-06 \\
    Spline Net & 8.32e-05 & 3.847e-03 & 1.07e-04 & 1.92e-05 & 7.492e-04 & 9.04e-04 \\
    Ansys & 1.46e-04 & 1.511e-02 & 1.072e-05 & 2.348e-05 & 8.912e-04 & 2.038e-02 \\
\end{tabular}
\caption{Quantitative results for implicit PINN, MAC grid, Spline Net and Ansys. Note: A direct comparison of these loss values is difficult as the losses had to be computed in different ways: The values for the implicit PINN and Spline Net were computed based on physics-informed losses while the losses for the MAC grid method were based on physics-constrained losses and the values of the Ansys-solver were computed based on the outputs on the underlying mesh representation. Thus, we argue that our results on drag and lift coefficients (see Table \ref{tab:cd_cl}) are more expressive in terms of overall performance.}
\label{tab:fluid_loss_quant}
\end{table*}

While the implicit PINN provides a very high accuracy for the momentum equation (see low $L_p$ values), it fails to achieve similar performance on matching the boundary conditions (see high $E[|\nabla\cdot \vec{v}|]$ values). The Marker-and-Cell grid based approach delivers the lowest loss values. However, these losses do not consider the inaccuracies delivered by the finite-difference approximations within the MAC grid itself. The high losses for the Ansys solver could stem from the fact that solutions are internally not optimized in strong form but based on Galerkin / Finite-Elements methods and that the provided derivatives might not be as accurate. 
We conclude that a comparison of such loss values is very difficult and one has to be very careful with drawing conclusions - especially on a Marker and Cell grid as the grid itself might introduce errors due to finite difference approximations. Furthermore, local loss values do not translate directly into physical accuracy for more global phenomena such as drag and lift forces.


\section{Acknowledgement}
This work has been supported by the German Research Foundation (DFG), FOR 2535 Anticipating Human Behavior.

\bibliographystyle{unsrt}  
\bibliography{aaai22}

\end{document}